\documentclass{article}

\usepackage[utf8]{inputenc} % allow utf-8 input
\usepackage[T1]{fontenc}    % use 8-bit T1 fonts
\usepackage{hyperref}       % hyperlinks
\usepackage{url}            % simple URL typesetting
\usepackage{booktabs}       % professional-quality tables
\usepackage{amsfonts}       % blackboard math symbols
\usepackage{nicefrac}       % compact symbols for 1/2, etc.
\usepackage{microtype}      % microtypography

\usepackage{systeme}
\usepackage{amsfonts}
\usepackage{amsmath}
\usepackage{amsthm}
\usepackage{amssymb}
\usepackage{algorithm}
\usepackage{algorithmic}
\usepackage{graphicx}
\usepackage{multirow}
\usepackage{wrapfig}
\usepackage{caption}
\usepackage{subcaption}
\usepackage[bottom]{footmisc}

\usepackage[final]{corl_2020} % Uncomment for the camera-ready ``final'' version.

\title{Evaluating the Safety of Deep Reinforcement Learning Models using Semi-Formal Verification}
\author{
  Davide Corsi*, Enrico Marchesini* and Alessandro Farinelli\\
  Department of Computer Science\\\
  University of Verona - 37134 Verona, Italy \\
  \texttt{*Contact authors: name.surname@univr.it} \\
}

\begin{document}
\maketitle

%\citet{Calandra2016}

%===============================================================================

\begin{abstract}
Groundbreaking successes have been achieved by Deep Reinforcement Learning (DRL) in solving practical decision-making problems. Robotics, in particular, can involve high-cost hardware and human interactions. Hence, scrupulous evaluations of trained models are required to avoid unsafe behaviours in the operational environment. However, designing metrics to measure the safety of a neural network is an open problem, since standard evaluation parameters (e.g., total reward) are not informative enough. In this paper, we present a semi-formal verification approach for decision-making tasks, based on interval analysis, that addresses the computational demanding of previous verification frameworks and design metrics to measure the safety of the models. Our method obtains comparable results over standard benchmarks with respect to formal verifiers, while drastically reducing the computation time. Moreover, our approach allows to efficiently evaluate safety properties for decision-making models in practical applications such as mapless navigation for mobile robots and trajectory generation for manipulators.
\end{abstract}

% Two or three meaningful keywords should be added here
\section{Introduction}
\label{sec:introduction}
	
	%\begin{wrapfigure}{b}{0.5\textwidth}
    %\centering
    %\includegraphics[width=0.5\textwidth]{Images/placeholder.png}
    %\label{fig:overview}
    %\caption{General Overview of the method.}
    %\end{wrapfigure}

Inspired by animal behaviour \cite{Sutton1998}, Reinforcement Learning (RL) aims at solving decision-making problems maximizing the expected cumulative reward, while an agent interacts with the surrounding environment to learn a policy. Challenging problems, however, are characterized by a high-dimensional search space that is not manageable by traditional RL algorithms. For this reason, Deep Reinforcement Learning gained attention in solving complex safety-critical tasks, due to its performance in a variety of areas, ranging from healthcare \cite{healtcare}, to robotics \cite{openai2019solving}. 

Despite recent impressive results, several factors limit the deployment of DRL models to control physical systems, outside of research demos. In tasks where high-cost equipment or humans are involved, the safety of a trained Deep Neural Network (DNN) must be evaluated to avoid potentially dangerous or undesirable situations. Furthermore, it is common practice to compare and evaluate DRL approaches in benchmarks that are considered standard (e.g., video games \cite{4} or simulated continuous locomotion \cite{38}), measuring their performance using target metrics such as total reward or number of successes in independent trials \cite{drlEval, drlEval2}. The usage of such metrics implies an underlying problem in the evaluation of the models that are collected during the training. Different sources of error (e.g., human-imperceptible perturbations in the input \cite{DC-3} or poor generalization \cite{drlEval}) are in fact challenging to detect with empirical testing phases. 

To address the safety problem of DNNs, a recent trend in DRL researches proposes the design of formal verification frameworks that, given a set of desired properties (e.g., a mobile robot does not turn in the direction of a close obstacle), either guarantee that the property is satisfied in a certain input space or return the input configurations that cause a violation \cite{J-0}. These methodologies aim at verifying whether the bound of a single output of a trained model lies in the desired interval (e.g., a motor velocity never exceeds a certain value) \cite{J-14, J-20} and the estimation accuracy of such output bounds have been successfully addressed by several recent studies \cite{J-44, DC-18}. The verification of DNNs that encode sequential decision-making policies, however, can not be directly addressed by previous formal approaches. Trivially, to verify these models, we should consider the relationships among multiple outputs. In a typical decision-making scenario, in fact, a network chooses the action that maximizes a return. Furthermore, as detailed in Section \ref{sec:preliminaries}, previous formal frameworks either are limited by poor scalability on high-dimensional DNNs, or require long wall-clock time to verify the designed safety properties.  
Against this background, in this paper, we propose a semi-formal verification approach (which we refer to as SFV), based on previous interval-analysis \cite{DC-x} formal verification approaches \cite{DC-16, DC-17}. We introduce two complementary metrics, \textit{safe rate} and \textit{violation rate}, to estimate the safety of trained models with respect to the desired properties. These values represent the percentage of the given input space that guarantees or violates a property (respectively), using limited computational demanding. The idea is to compute the output bounds using a semi-formal interval propagation method, detailed in Section \ref{sec:methods:semiformal}. Moreover, this allows to obtain comparable results in terms of safe and violation rate while drastically reducing the computation time, with respect to formal verifiers. We designed SFV to provide an evaluation of the behavior of a trained agent, finding the best performing model among all the trained ones that, at a first glance (i.e., comparing standard metrics such as total reward), seems comparable in terms of performance. Crucially, SFV allows to evaluate the high-dimensional (continuous) input space for all the trained models, which is either not feasible by simulation or very time consuming with formal verifiers (that are also not directly applicable in decision-making problems).
We empirically evaluate SFV on a standard benchmark, CartPole \cite{DC-30}, and on two decision-making task of real interest in recent DRL literature: (i) mapless navigation \cite{50, 59} for a TurtleBot3 (a widely considered platform in several works \cite{1, 48}), and (ii) trajectory generation for a commercial manipulator  \cite{3, 33} (among the variety of platforms, we considered a manipulator for their wide utilization in industry).

Summarizing, this work makes the following contributions: (i) we introduce SFV, a semi-formal verification approach that enables to estimate the safety of trained DNNs that encodes real decision-making tasks; (ii) our evaluation in Section \ref{sec:methods:semiformal} shows that the output bounds computation, based on novel semi-formal interval propagation allows to drastically reduce the verification time while obtaining comparable results with respect to a state-of-the-art approach \cite{DC-17}, on the ACAS models \cite{DC-32}, a standard benchmark to compare formal verification tools, and (iii) Section \ref{sec:evaluation} shows that standard metrics (e.g., total reward) are not informative enough to evaluate a model, and we introduce safe and violation rates to measure the reliability of a network with respect to designed safety properties.
\section{Preliminaries and Related Work}
\label{sec:preliminaries}

Significant effort has been recently devoted to safety in DRL, which can be addressed with a variety of methods. A wide branch of literature proposes the introduction of constraints in the exploration phase to limit the learned behavior of the agent, or the use of well-designed reward functions to encourage or discourage certain actions \cite{DC-1, DC-2}. These methods, however, aim at minimizing undesirable behaviors without providing any formal guarantees. Trivially, the input space cardinality is infinite so it is not possible to test or simulate all the configurations to ensure their safety. DNNs, in fact, are vulnerable to adversarial attacks and can suffer from a poor generalization to unseen states \cite{J-29}.
In contrast, the formal verification of DNNs \cite{J-0} that do not encode decision-making problems has been addressed exploiting Boolean Satisfiability or Satisfiability Modulo Theories (SMT), searching for configurations that falsify an assertion \cite{DC-9, J-7}. A different approach exploits optimization techniques \cite{J-4} for such search \cite{J-23, J-38, J-31}. These methods, however, are strongly limited by their scalability on large networks and require a significant amount of computation, in particular for non-linear constraint networks.
Recently, several approaches aim at searching input configurations that deny a given safety property, using a layer-by-layer analysis to partially overcome the limitations of previous approaches. An example is FastLin \cite{J-41} which combines search and reachability to compute and analyze the absolute bounds of an output. Another group of methods such as Neurify \cite{DC-17} (an improvement over ReluVal \cite{DC-16}), DeepPoly \cite{DC-18} and others \cite{J-20, J-13}, rely on an accurate bound computation to propagate the inputs and calculate the corresponding output bounds, using interval analysis. We also exploit this analysis, which is briefly described in the next section.

\subsection{Interval Analysis}

\begin{wrapfigure}{b}{0.5\textwidth}
        \centering
        \includegraphics[width=0.5\textwidth]{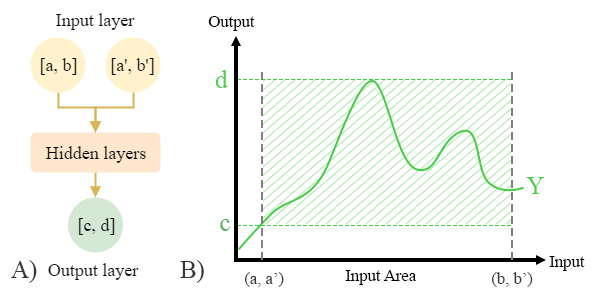}
        \caption{Example of a bound analysis of a generic neural network with 2 inputs and 1 output.}
        \label{fig:bound_overall}
    \end{wrapfigure}

In this section, we introduce the main concepts and notations adopted throughout the paper.
We define an \textit{area} as a set of inputs, limited by an upper and lower bound. A \textit{subarea} is a further subdivision of the same input set. 
Figure \ref{fig:bound_overall} shows an example of these concepts, where area = $([a, b], [a', b'])$ (i.e., the input area to evaluate). 
Two possible subareas of this are $([a, (a + b)/2], [a', b'])$ and $([(a + b)/2, b], [a', b'])$, notice that the union of the subareas, always returns the former area.
Figure \ref{fig:bound_overall} also shows an example of two different bound representations: (i) a simple network with two inputs and one output (Figure \ref{fig:bound_overall}A); (ii) a graphical overview (Figure \ref{fig:bound_overall}B), where the area is bounded by values [a, b] on the first input and by [a', b'] on the second one. 
The $Y$ curve is the representation of the network output $f(X)$ and [c, d] is the bound for the values that the function can assume.

Summarizing, we visualize the bound analysis of a DNN as a 2-dimensional graph with the inputs of the network (i.e., $x \in X$) on the x-axis and the output values (i.e., $f(X)$) on the y-axis. To visualize the bounds when $n > 1$ (where $n$ is the number of input nodes), we assume that each point on the x-axis represents a tuple of $n$ values. Notice that, the ordering among the tuples on the x-axis is arbitrarily defined and is required only for visualization purposes (i.e., it does not affect the analysis).

\subsection{Safety Properties in Decision-Making Problems}
\label{sec:preliminaries:formal}
Following the standard representation provided by \citet{J-0}, a safety property can be formalized in the following form (where $x_k \in X$, with $k \in [0, n]$ and $y_j$ is a generic output):

\begin{equation} \label{equation_1}
\Theta: \mbox{If } x_0\in[a_0, b_0] \land ... \land x_n\in[a_n, b_n] \Rightarrow y_j\in[c, d]
\end{equation}

In this work, however, we focus on neural networks that encode decision-making problems, where each output node represents an action, and the action corresponding to the node with the highest value is usually chosen. In this scenario, we are not interested in the absolute value that each output node assumes. In contrast, we aim at verifying whether the value of an output is lower (or greater) than the value of another one. For this reason, we reformulate Proposition \ref{equation_1} in the following form:

\begin{equation} \label{equation_2}
\Theta: \mbox{If } x_0\in[a_0, b_0] \land ... \land x_n\in[a_n, b_n] \Rightarrow y_i < y_j
\end{equation}

where the relation between $y_i$ and $y_j$ can be verified using interval algebra of \citet{DC-x}. In particular, supposing $y_i= [a, b]$ and $y_j=[c, d]$ we obtain the proposition:
\begin{equation} \label{equation_3} 
b < c  \Rightarrow y_i < y_j
\end{equation}

\begin{figure*}[t]
   \centering
	\includegraphics[width=0.85\linewidth]{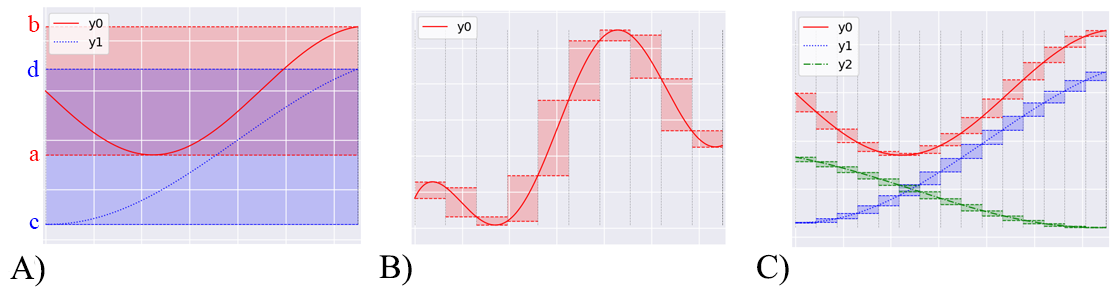}
 	\caption{Explanatory output analysis of: (A) decision-making problem with two outputs and one subdivision; (B) estimation of an output function shape, using multiple subdivisions; (C) output analysis with three outputs and multiple subdivisions.}
  	\label{fig:interval_analysis}
\end{figure*}

Here we describe how to exploit the iterative refinement proposed by \citet{DC-16}, that reduces the overestimation of the output bound computation, to also directly verify safety properties for decision-making scenarios. Afterward, Section \ref{sec:methods:semiformal} introduces the lightweight semi-formal approach used to estimate our novel metrics. 
Previous approaches exploit a formal input area layer-by-layer propagation (we refer the interested reader to the original implementation for more details \cite{J-46}), to obtain a strict estimation of the output bounds. However, such technique returns bounds that are not informative for a decision-making problem. Even with a perfect estimation of the output bounds\footnote{The real shape of the two output function is impossible to obtain due to the non-linearity of DNNs and here we show two explanatory curves}, Figure \ref{fig:interval_analysis}A shows that if $y_1$ $<$ $y_0$ (i.e., two output functions) in the given input area, $d \nless a$ so, with Proposition \ref{equation_3}, we can not formally conclude if the decision-making property is proved or denied. 
To address this, we exploit the \textit{iterative refinement} to also obtain an estimation of the output curves (the overestimation reduction provided by this method is also important to make the separation between the outputs function more clear). Figure \ref{fig:interval_analysis}B shows this key process, where increasing the number of subdivisions we compute an arbitrary precise estimation of the real function. Figure \ref{fig:interval_analysis}C shows how we perform the evaluation, where for each generated subarea we check if the property is respected. In order to formally verify a property, Proposition \ref{equation_2} has to hold for all the subareas; while to deny a property we just need one counterexample. For these reasons, with a sufficient number of subdivisions, we can use an approximate shape of the output functions, to always assert if a property is respected or not.

\subsection{Subareas Analysis}
\begin{figure*}[t]
   \centering
	\includegraphics[width=1\linewidth]{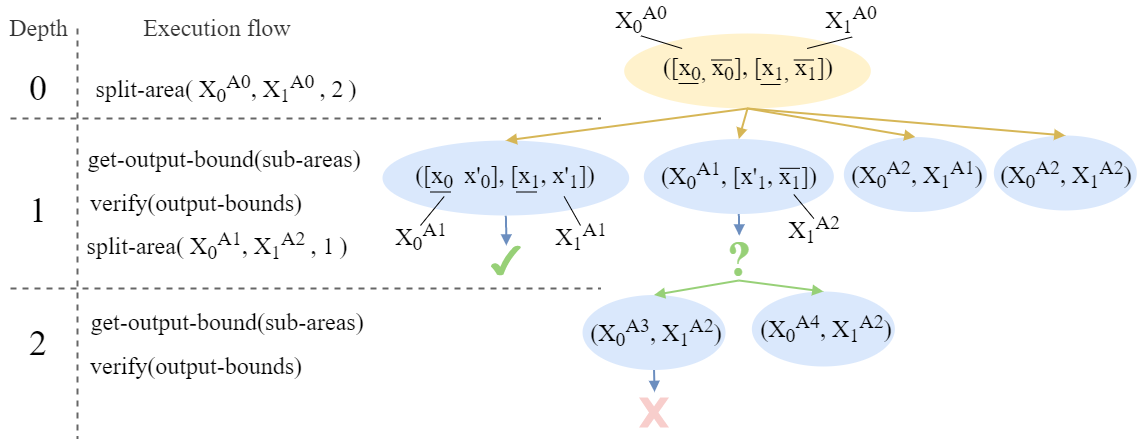}
 	\caption{Explanatory example of the iterative bisection tree generated to verify the property $y_0 < y_1$ in the given input area ($X_{0}^{A_0}$, $X_{1}^{A_0}$). As example, $X_{1}^{A_1}$ and $X_{1}^{A_2}$ derive from the split of the original input area $X_{1}^{A_0}$, where $x_{1}'$ indicates the new value that limits that range.}
  	\label{fig:areatree}
\end{figure*}

In the previous section, we showed that with an appropriate number of subdivisions we can check a safety property in the form of Proposition \ref{equation_2}. However, it is not possible to estimate the exact number of the necessary subdivision a priori. For this reason, we encode our approach as a tree search problem, that iteratively subdivides the input area (or subareas) until it reaches a solution (i.e., the scenario in Figure \ref{fig:interval_analysis}C). In fact, we further subdivide a subarea when we can not prove (or deny) a property, hence, we need a smaller input area (Figure \ref{fig:interval_analysis}A).
Figure \ref{fig:areatree} shows an example of a possible execution flow on a network with two inputs (i.e. $x_0$ and $x_1$).
Initially, the root of our tree, to which we refer as \textit{subarea tree}, contains the input areas ($X_{0}^{A_0}$ and $X_{1}^{A_0}$ in Figure \ref{fig:areatree}, where $X_{i}^{A_j}$ represents a network input range with lower bound $\underline{x_{i}}$ and upper bound $\overline{x_{i}}$). 
This example considers a split of the input area in $2$ sections, following a random strategy\footnote{It is possible to use different heuristics to optimize the search. However, in our experiments, the usage of a different strategy, i.e., split the biggest area first, does not provide a significant improvement over the random strategy.}. We assume to split first $X_{0}^{A0}$ obtaining $X_{0}^{A_1}, X_{0}^{A_2}$, and then $X_{1}^{A0}$, obtaining $X_{1}^{A_1}, X_{1}^{A_2}$, where $x_{i}'$ represents a new input range limit (e.g., one subdivision of the initial $[\underline{x_{i}}, \overline{x_{i}}]$, generates two intervals $[\underline{x_{i}}, x_{i}']$ and $[x_{i}', \overline{x_{i}}]$). The combinations of all the new bounds represents the next layer of the tree, which is the \textit{unverified-subarea-tree} (Figure \ref{fig:areatree} at depth 1). Hence, in this example, the first subarea $[X_{0}^{A_1}, X_{1}^{A_1}]$ shows the situation in Figure \ref{fig:interval_analysis}C (where the property $y_1 < y_0$ holds for each subarea) and the algorithm continues with the verification of the next subarea $[X_{0}^{A_1}, X_{1}^{A_2}]$. Here the example falls into the case of Figure \ref{fig:interval_analysis}A and further subdivisions are required to verify the property in this branch (Figure \ref{fig:areatree} at depth 2). Finally, $[X_{0}^{A_3}, X_{1}^{A_2}]$ represents the case in Figure \ref{fig:interval_analysis}C, where the property $y1 < y_2$ is violated in the subareas after the intersections of the two curves.
 
\section{Evaluation Methods}
\label{sec:methods}

In this section, we present how SFV evaluates the safety metrics of neural network over a set of desired properties, in decision-making tasks. We start by introducing two complementary metrics to quantify the safety of a trained model and describe the semi-formal interval propagation method.

\subsection{Safe and Violation Rates}
\label{sec:methods:saferate}
Standard performance metrics used in DRL to evaluate a model typically includes average total reward or some derivations such as the number of successes in independent trials. However, given a set of models that achieve similar reward, we can not directly infer how they behave with respect to some desired safety properties. For this reason, we aim at measuring the safety over such properties, introducing a novel evaluation metrics. Section \ref{sec:evaluation} shows that, in practical applications (e.g., robotics), it provides a more reliable estimation on the safety of the model.

We introduce the \textit{violation rate} as a the percentage of the input area that causes a property violation. Hence, it is the normalized sum, with respect to the input area dimension, of the subarea sizes that presents a violation (e.g. the subarea $[X_{0}^{A_3}, X_{1}^{A_2}]$ in Figure \ref{fig:areatree}). When considering multiple properties in a single task (as in Section \ref{sec:evaluation}), we compute the violation rate as the average of the violation rates for the different properties. Conversely, we define the \textit{safe rate} as the percentage of the input area that respect the properties (e.g. the subarea $[X_{0}^{A_1}, X_{1}^{A_1}]$ in Figure \ref{fig:areatree}).

Crucially, these novel rates do not measure the true probability of a property violation in typical task execution, as it usually represents an upper or lower bound. In fact, within an input area, the different state configurations do not have the same probability to appear (and the metrics do not consider this). 
The heterogeneous distribution in a high-dimensional state space, makes not possible to compute an accurate value for these novel metrics using standard empirical evaluations (as shown in Section \ref{sec:evaluation}), which typically return a very imprecise safe rate. Hence, the idea is to efficiently provide a strict estimation of the safe rate using our semi-formal method on the multitude of trained models that achieve comparable performance (e.g., total reward). 

\subsection{Semi-Formal Interval Propagation}
\label{sec:methods:semiformal}
Current state-of-the-art approaches require high computational demanding that limits their application as model evaluators. Furthermore, they can not be directly applied to decision-making problems. 
In this section, we present our semi-formal interval propagation method to estimate the safe rate, while drastically reducing the computational demanding (i.e., in terms of time) of the verification phase. Section \ref{sec:methods:comparison} shows that the semi-formal implementation obtains comparable safe rates with respect to a formal verification approach in a benchmark problem.

%\begin{figure}[t]
%   \centering
%    \includegraphics[width=0.32\linewidth]{Images/method_scalability.png}
%    \includegraphics[width=0.32\linewidth]{Images/overestimation.png}
%	\includegraphics[width=0.32\linewidth]{Images/underestimation.png}
%	\caption{Placeholder A) and B) ...}
%  	\label{fig:semi-formal}
%\end{figure} 

In more detail, formal methods based on bound analysis suffer from the following issues that servery limits their efficiency: (i) the \textit{overestimation} and (ii) the \textit{propagation time} (especially on high-dimension networks). The former is partially solved by the iterative refinement (detailed in Section \ref{sec:preliminaries}), however, it requires a very small input area to compute a strict estimation of the bounds. The latter, in contrast, can not be solved as it is an intrinsic problem of formal propagation methods.

To address these issues, we introduce a semi-formal propagation approach, based on an empirical evaluation of the output bounds. In detail, we sample $n$ points from the interested subarea and compute the corresponding $n$ values for each output node, using a simple forward step of the network. Afterward, we estimate the output bound, using the maximum and the minimum value obtained for each node. Ideally, if $n \rightarrow \infty$, we obtain an exact estimation of the output bounds. Crucially, the next section shows that we do not require big values for $n$, to obtain a good estimation.

\subsubsection{Comparison with Formal Verification}
\label{sec:methods:comparison}
\begin{figure}[t]	
	\centering
	\begin{subfigure}[t]{0.30\linewidth}
		\centering
		\includegraphics[width=1\linewidth]{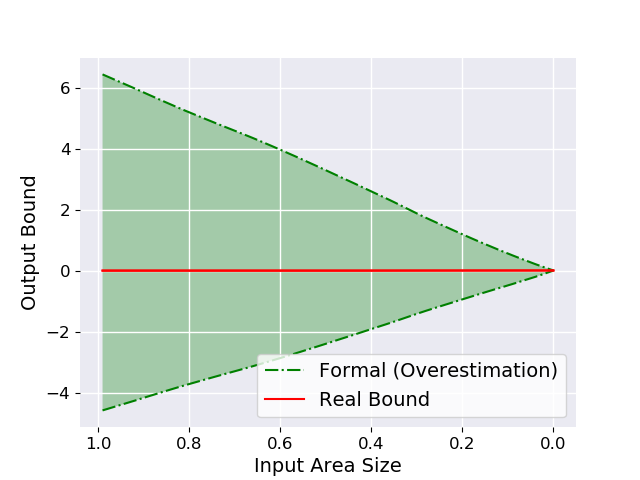}
		\caption{Overestimation }
		\label{fig:semi-formal:a}
	\end{subfigure}
	\quad
	\begin{subfigure}[t]{0.30\linewidth}
		\centering
		\includegraphics[width=1\linewidth]{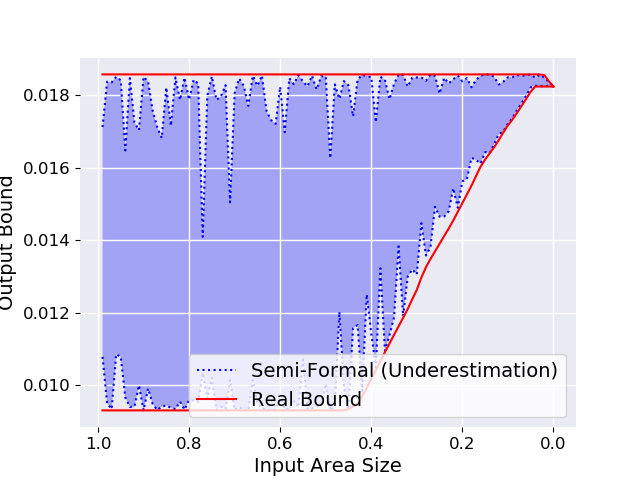}
		\caption{Underestimation}
		\label{fig:semi-formal:b}
	\end{subfigure}
	\quad
	\begin{subfigure}[t]{0.30\linewidth}
		\centering
		\includegraphics[width=1\linewidth]{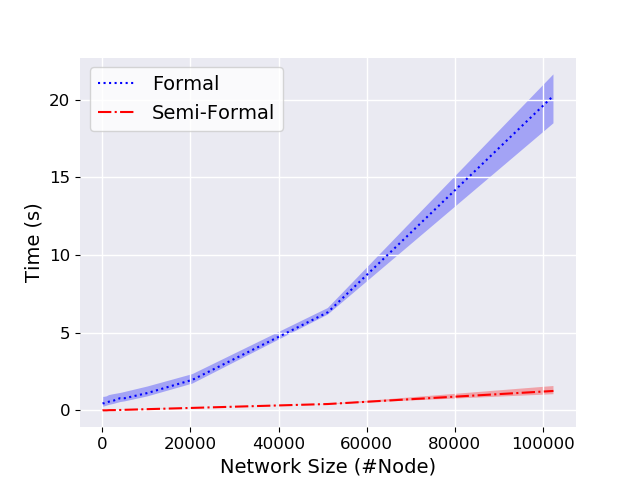}
		\caption{Scalability}
		\label{fig:semi-formal:c}		
	\end{subfigure}
	\caption{Comparison between Neurify and our semi-formal method with $n=20$. Red lines represent the \textit{"real" bounds} of the input area.}
	%The \textit{real bounds} is a strict estimation, obtained with a test of more than 100k samples.}
	\label{fig:semi-formal}
\end{figure}

Figure \ref{fig:semi-formal} shows the output bound computation results of our method compared with Neurify \cite{DC-17}, a state-of-the-art formal verification tool. 
In detail, Figure \ref{fig:semi-formal:a} shows the overestimation problem where formal methods present a high overestimation when computing the output bounds on big input areas. To address this, they require multiple executions of the iterative refinement to obtain smaller areas and compute a strict estimation of the output bounds, causing exponential growth in time.
In contrast, Figure \ref{fig:semi-formal:b} shows that our semi-formal approach does not suffer from overestimation on big input areas as it uses real input configurations for the estimation. Hence, there is a clear limitation of SFV as it presents an implicit underestimation on the output bounds. Considering the entire input area (i.e., the normalized value $1$ in Figure \ref{fig:semi-formal}) with a "real"\footnote{To compute a significant estimation of the real output bound, we run the semi-formal propagation with $n$ equals to ten million configurations} output bound of size $0.0091$, Neurify overestimates it in a size of $11$, while SFV underestimates it in a size of $0.0063$. Summarizing, SFV obtains a tight underestimation of the output bounds regarding the size of the input area, while Neurify requires smaller areas to compute a precise estimation. However, to obtain a formal verification, it is not acceptable to use an underestimation of the bounds as it can lose information; SFV, in contrast, aims at computing an estimation of the safety metrics as an indicator of the behaviour of the trained models.
Moreover, Figure \ref{fig:semi-formal:c} shows how the computational demanding of the formal propagation compared with semi-formal one, varies with respect to the size of the network. In particular, it highlights that the semi-formal approach drastically scales better on the network size.

\begin{figure}[b]
   \centering
    \includegraphics[width=0.30\linewidth]{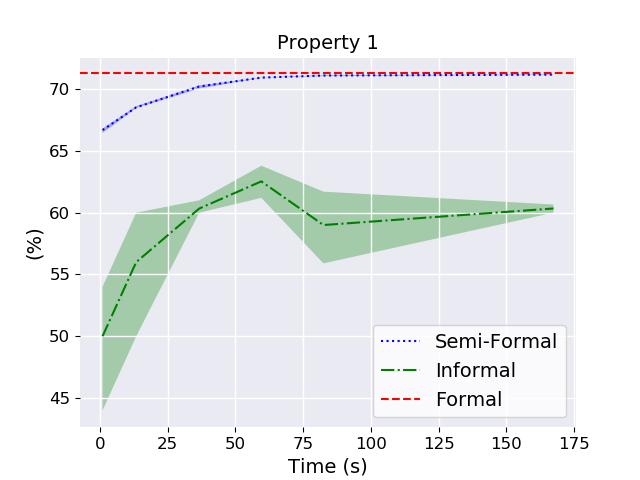}
	\includegraphics[width=0.30\linewidth]{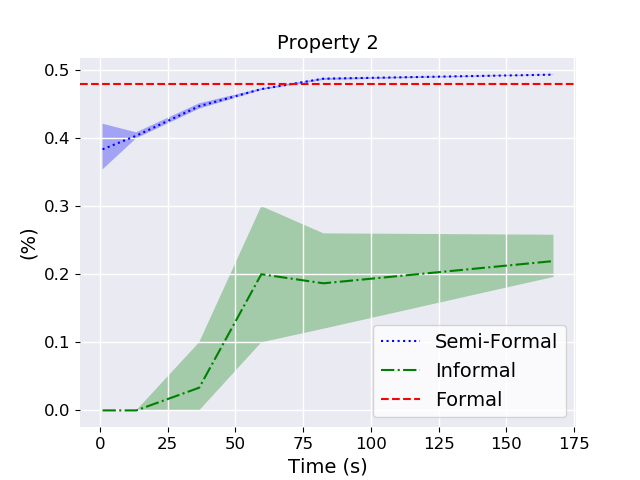}
	\includegraphics[width=0.30\linewidth]{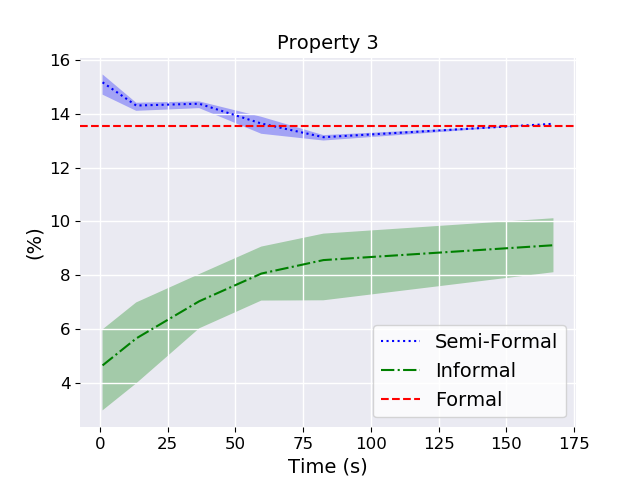}
 	\caption{Comparison between SFV, the informal one and the formal \textit{safe rate} computed with Neurify, on three properties of the standard ACAS benchmark. Neurify requires 363.449s for the property 1, 240.07s for the property 2 and 274.341s for the property 3.}
  	\label{fig:formal-comparison}
\end{figure}

To further validate the safe rate estimation of our method, we compare Neurify and SFV on a standard verification benchmark, the ACAS models \cite{DC-32}. 
Figure \ref{fig:formal-comparison} shows the comparison between Neurify, SFV, and an informal method, which computes the safe rate using only simulation (i.e., we run the models to collect how many propertiy failures they encounter). Crucially, our semi-formal method returns comparable results with respect to the formal verifier with limited computation time. In detail, to obtain a comparable safe rate (i.e., with error $< 1\%$), SFV is, $726\%$, $400\%$ and $456\%$ (i.e., $527\%$  on average) faster than Neurify in the considered ACAS properties.
In the next section, we evaluate SFV in three different decision-making tasks, showing that the violation (or safe) rate is a valuable metric to measure safety in DRL applications.

\section{Empirical Result}
\label{sec:evaluation}

Our goal is to measure the reliability of the trained models in decision-making applications, comparing how typical evaluation metrics such as total reward and success rate differ from the proposed safe rate. We aim to spur discussion about how to interpret the behaviour of different models, to choose the safest and best performing one.
In detail, we show that trained networks that present high scores with standard metrics, actually present very different behaviours in practice (i.e., models with comparable rewards have very different safety scores) that can lead to undesirable situations. We empirically evaluate our framework on a well-known DRL benchmark, CartPole \cite{DC-30}, and two real robotic applications that are widely considered in recent DRL literature: mapless navigation \cite{1, mapless1, 59} and trajectory generation for a commercial manipulator \cite{3, 33}. The considered agents are modeled using two relu hidden layers with 64 neurons each and trained using the Rainbow algorithm \cite{68} (we refer the interested reader to the original papers for further details about training algorithm and hyperparameters).

\paragraph{Cart Pole}
The OpenAI Gym \cite{38} problem is solved when collects an average reward of 195.0 over 100 consecutive trials, according to the original version of CartPole \cite{BartoCartPole}.
When the pole is offset by $> 15$ $deg$ from the vertical position, or the cart is $> 2.4$ units from the center, an episode ends. Following official documentation, the network has inputs: (i) cart position $x_0 \in [-4.8, 4.8]$; (ii) cart velocity $x_1 \in [-Inf, Inf]$; (iii) pole angle $x_2 \in [-0.24, 0.24]$, (iv) pole velocity $x_3 \in [-Inf, Inf]$ and discrete outputs: (i) push cart to left $y_0$ and (ii) push cart to right $y_1$. According to these limits, we formalize two safety properties, with values normalized in range [0, 1], as follows:
$
\textbf{$\Theta_{C, 0}$}: \mbox{If } 
x\textsubscript{0} \in[0.2, 0.8] \land x\textsubscript{1}\in[0.4, 0.6] \land
x\textsubscript{2}\in(0.7, 1] \land
x\textsubscript{3}\in[0.5, 1] \Rightarrow y\textsubscript{0} < y\textsubscript{1}$

$
\textbf{$\Theta_{C, 1}$}: \mbox{If } 
x\textsubscript{0} \in[0.2, 0.8] \land x\textsubscript{1}\in[0.4, 0.6] \land
x\textsubscript{2}\in[0, 0.3) \land
x\textsubscript{3}\in[0, 0.5] \Rightarrow y\textsubscript{1} < y\textsubscript{0}$

These properties aim at verifying that when the pole reaches its angle limit, the cart pushes in the opposite direction to move the pole towards the vertical alignment.

\paragraph{Mapless Navigation}
In this navigation task, a robot must reach a target using local observation to avoid obstacles and without a map of the environment. Our problem formalization is similar to the one presented in \cite{mapless1}, using a Turtlebot3 with constant linear velocity. The network has 21 inputs: (i) 19 sparse scan values $x_0, .., x_{18}$ normalized between $[0,1]$, sampled in a fixed angle distribution between -90 and 90 $deg$; (ii) target polar coordinates with respect to the robot (i.e., heading $x_{19}$ and distance $x_{20}$, normalized in $[0, 1]$); and three outputs for the angular velocities (i.e., [-90, 0, 90] deg/s). 
In this task, we evaluate the safety of the network and compare it with the success rate (i.e., how many correct trajectories are performed in the last 100 epochs) using two safety properties, which can be described in natural language as:

\noindent\textbf{$\Theta_{T, 0}$:} If there is an obstacle too close to the left and one in front, whatever the target is, turn right. \\
\noindent\textbf{$\Theta_{T, 1}$:} If there is an obstacle too close to the right and one in front, whatever the target is, turn left.

\paragraph{Trajectory Generation}
In this scenario, the agent has to rotate the joints to generate a real-time trajectory to reach a target. The formalization of this problem is similar to the one presented in \cite{33}, where the input layer contains 9 nodes normalized in range $[0, 1]$: (i) one for each considered joint; (ii) three for target coordinate; and 12 nodes for the output: each joint is represented by 2 nodes to move it $\omega$ degrees clockwise or anti-clockwise. This encoding of the output allows a straightforward verification process for our tool (i.e., one node represents only one specific action).
Hence, we designed safety properties to check if the manipulator operates inside its work-space, considering properties in the following form: if the current angle of a joint $j_i$ equals to its domain limits (left or right), the robot must not rotate $j_i$ in the wrong direction (i.e. an action that rotates $j_i$ causes the robot to exit from the work-space). This translates in the following formalization:

$
\textbf{$\Theta_{P, 0L}:$} \mbox{If } x_0\in[1, 1] \land x_1, ..., x_8 \in D \Rightarrow y_0 < [y_1, ..., y_{11}], \textit{where D = (0, 1)}
$

$
\textbf{$\Theta_{P, 0R}$:} \mbox{If } x_0\in[0, 0] \land x_1, ..., x_8 \in D \Rightarrow y_1 < [y_0, y_2, ..., y_{11}], \textit{where D = (0, 1)}
$

where $\Theta_{P, 0L}$ is a configuration where $j_0$'s angle equals to its limit on the left (i.e., a normalized value 1). The output value corresponding to the action \textit{rotate left}, must be lower than at least one of the others. For each joint $j_i$ we consider two properties, one for the left limit ($\Theta_{P, iL}$) and one for the right limit ($\Theta_{P, iR}$).

\subsection{Results}

\begin{figure}[t]	
	\centering
	\begin{subfigure}[t]{0.30\linewidth}
		\centering
		\includegraphics[width=1\linewidth]{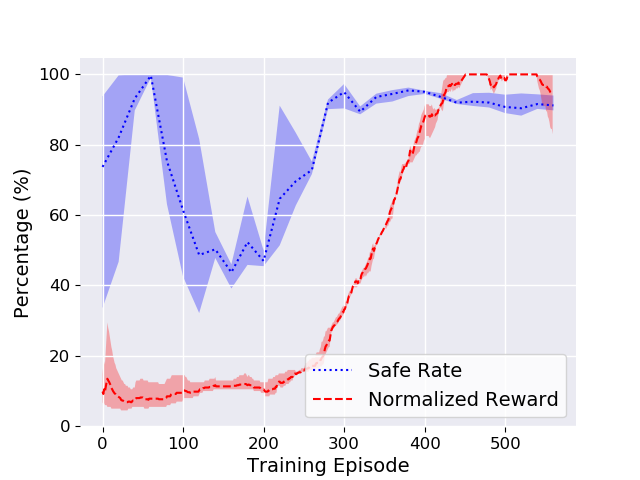}
		\caption{CartPole}
		\label{fig:final-results:a}		
	\end{subfigure}
	\quad
	\begin{subfigure}[t]{0.30\linewidth}
		\centering
		\includegraphics[width=1\linewidth]{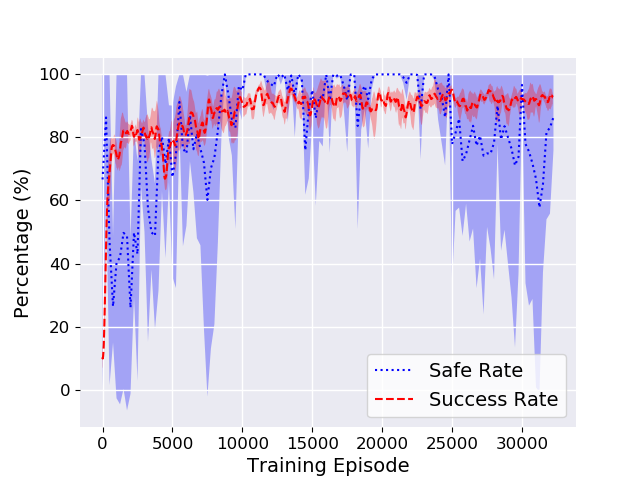}
		\caption{Mapless Navigation }
		\label{fig:final-results:b}
	\end{subfigure}
	\quad
	\begin{subfigure}[t]{0.30\linewidth}
		\centering
		\includegraphics[width=1\linewidth]{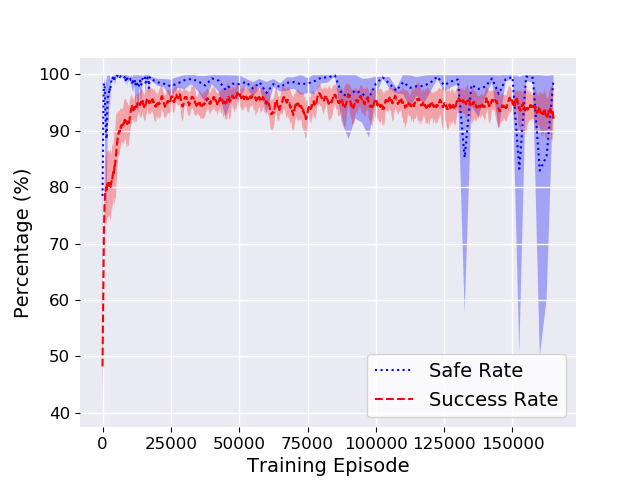}
		\caption{Robotic Manipulator}
		\label{fig:final-results:c}
	\end{subfigure}
	\caption{Comparison between the \textit{normalized reward} (or \textit{success rate}) and the \textit{safe rate}.}
	\label{fig:final-results}
\end{figure}

In order to collect statistically significant data, we performed five training phases for each task, using different random seeds \cite{colas2019hitchhikers}. We performed our experiments on an i7-9700k and a GTX2070, using the implementation described in Section \ref{sec:methods}. For each graph, the curves report the average and the standard deviation of the runs, considering: (i) standard metrics (i.e., normalized reward or success rate), where the success rate is how many successful trajectories are performed in the case of the TurtleBot3 and the manipulator; (ii) safe rate: the average over the safe rates (Section \ref{sec:methods:saferate}) for each considered property. These metrics are smoothed on one hundred episodes (expect for CartPole, where results are smoothed on ten episodes).

Figure \ref{fig:final-results} shows that, in our experiments, the safe rate is always characterized by high values (i.e., properties hold in the input areas) in the early stage of the training phases. We motivate this as the models initially choose random actions to explore the state space, while collecting reward to infer the task. A more detailed analysis shows that all the agents tend to stay still or move close to their initial position. Afterward, the safe rate starts to follow a similar trend to the considered standard metrics as the agent is learning the required behaviour to solve the task. 
In contrast, in the advanced stages of the training, when agents successfully address the task, the safe rate curve behaves differently. Figure \ref{fig:final-results:a} refers to the CartPole environment, where the safe rate has a similar trend to the normalized reward as a poor generalization is required to solve the problem. Mapless navigation in Figure \ref{fig:final-results:b} has a drop in the safe rate after many successful epochs where, according to the success rate, the agent seems to solve the task. This evidence the problem described in the previous sections as the multitude of trained models with similar standard performance, actually have very different behaviors. In particular, we noticed that around epoch 25000, the network tries to learn shorter paths by navigating closer to obstacles, leading to more unsafe behaviours. Figure \ref{fig:final-results:c} shows similar results for the trajectory generation task that behaves similarly to the navigation problem.
Summarizing, our experiments show that our novel safe (or violation) rate can evaluate the behavior of the trained models with respect to the designed safety properties. This allows to find the best performing model among all the trained ones that seem comparable in terms of performance using standard metrics.
\section{Conclusion}
\label{sec:conclusion}

We present SFV, an efficient novel framework designed as a semi-formal verification tool for the analysis of safety properties for real-world decision-making problems. We compare its performance with respect to previous formal verification tools on the ACAS benchmark. Moreover, we evaluate SFV on the CartPole problem, and on two robotic scenarios of real interest in DRL literature, mapless navigation, and trajectory generation. 

Our semi-formal interval propagation method for the output bounds computation, allows SFV to drastically outperform existing verification approaches in terms of time while obtaining comparable safety results. Moreover, we introduce two complementary metrics to measure the reliability of a trained model, showing that standard metrics such as total reward are not enough informative to evaluate the behaviour of the model. This paper paves the way for several important research directions which include exploiting the computational efficiency of SFV to evaluate models in the training phase, to guarantee that the network maximizes the expected cumulative reward while resulting in safer trained models.

%===============================================================================

% The maximum paper length is 8 pages excluding references and acknowledgements, and 10 pages including references and acknowledgements

\clearpage
% The acknowledgments are automatically included only in the final version of the paper.
%\acknowledgments{If a paper is accepted, the final camera-ready version will (and probably should) include acknowledgments. All acknowledgments go at the end of the paper, including thanks to reviewers who gave useful comments, to colleagues who contributed to the ideas, and to funding agencies and corporate sponsors that provided financial support.}

%===============================================================================

% no \bibliographystyle is required, since the corl style is automatically used.
\bibliography{biblio}  % .bib

\end{document}